\documentclass[11pt]{article}

\usepackage[preprint]{acl}

\usepackage{times}
\usepackage{latexsym}
\usepackage{tcolorbox}
\tcbuselibrary{breakable}
\usepackage[T1]{fontenc}

\usepackage[utf8]{inputenc}

\usepackage{microtype}

\usepackage{inconsolata}

\usepackage{graphicx}
\usepackage{amsmath}

\usepackage{algorithm}
\usepackage{algpseudocode}
\usepackage{amssymb}
\usepackage{enumitem}

\usepackage{float}
\usepackage{color}

\tcbuselibrary{skins} 
\definecolor{lightgray}{gray}{0.95}

\usepackage{booktabs}
\usepackage{bm}

\title{STEVE: Stabilizing Textual Gradient-Based Prompt Optimization via Error-Driven Refinement and Regularized Verification}

\author{
  \textbf{Yifan Xu}\textsuperscript{1}\thanks{Corresponding author: \href{mailto:yx52@illinois.edu}{yx52@illinois.edu}} \quad
  \textbf{Yixuan Li\textsuperscript{1}} \quad
  \textbf{Xinzhuo Li\textsuperscript{1}} \quad
  \textbf{Yixin Gu\textsuperscript{1}}
  \\
  \textbf{Yifan Shen\textsuperscript{1}} \quad
  \textbf{Lijun Yu\textsuperscript{2}} \quad
  \textbf{Haohan Wang\textsuperscript{1}}
  \\
  \textsuperscript{1}University of Illinois Urbana-Champaign \quad
  \textsuperscript{2}Google DeepMind
}

\usepackage{makecell}
\usepackage{multirow}
\usepackage{caption}
\definecolor{mygreen}{HTML}{34A853}
\definecolor{myred}{HTML}{EA4335}

\newcommand{\gain}[1]{\textcolor{mygreen}{\tiny{$\uparrow$}#1}}
\newcommand{\loss}[1]{\textcolor{myred}{\tiny{$\downarrow$}#1}}
\ifdefined\pdfobjcompresslevel
\fi

\begin{document}
\maketitle

\begin{abstract}
Textual-gradient methods automate prompt optimization through natural-language feedback, but their iterative updates can be unstable. We identify two sources of this instability: noisy gradients produced from already-correct examples and over-specialization to hard cases that degrades performance on simpler inputs. We introduce STEVE, a stabilization framework with two coupled mechanisms. Error-Driven Refinement generates gradients only from incorrectly handled examples, concentrating updates on informative failures. Regularized Verification treats every update as provisional and accepts it only when improvement on hard cases does not cause unacceptable regression on a preservation set. Across ten reasoning benchmarks, three evaluator/optimizer models, and established prompt-optimization baselines, STEVE reduces degradation and produces more robust prompts. Additional evaluations with \texttt{gpt-5.4-mini}/\texttt{gpt-5.4} on symbolic reasoning, GSM8K-Platinum, and DS-1000 show that these gains persist with newer models and larger test sets. STEVE therefore provides a practical way to improve the stability and effectiveness of textual-gradient prompt optimization.
\end{abstract}

\section{Introduction}
\label{sec:introduction}

\begin{figure*}[htbp]
    \vspace{-0.2cm}
    \centering
    \includegraphics[width=0.97\textwidth]{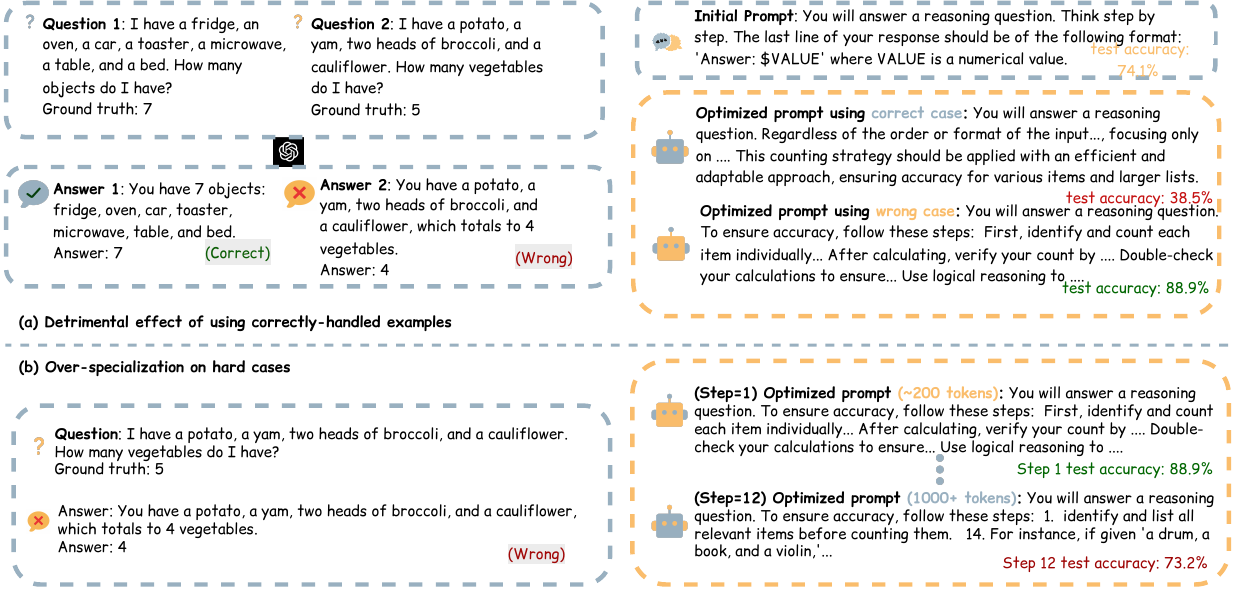}
    \vspace{-0.3cm}
    \caption{(a) \textbf{Feedback from correct versus failed cases}. Using a prompt with 74.1\% baseline accuracy, the executor answers Question 1 correctly but fails on Question 2. Updating from the failed case improves test accuracy to 88.9\%, whereas updating from the correct case lowers it to 38.5\%. (b) \textbf{Over-specialization}. Repeated refinement without verification initially raises accuracy to 88.9\% but eventually produces a 1000+-token prompt whose accuracy falls to 73.2\% after 12 steps.}
    \vspace{-0.1cm}
    \label{fig:error}
\end{figure*}

Large Language Models (LLMs) have demonstrated remarkable capabilities across a wide range of applications, from complex reasoning \citep{zhang2025ratt, yao2023react, kojima2022large} to acting as the backbone for autonomous agents \citep{wang2024survey, schick2023toolformer, park2023generative}, including agents that interact with graphical interfaces \citep{liu2026augmenting}. However, their performance is sensitive to the input prompt. More broadly, the composition of contextual information can systematically alter LLM behavior, underscoring the need to optimize and control prompts carefully \citep{wang2025context}. Manually engineering prompts to elicit optimal performance is a tedious process that is often unpredictable and difficult to scale. While automated prompt optimization (APO) methods \citep{zhang2024sprig, 11106067} offer a promising alternative, they face challenges due to the stochasticity of the model's outputs. First, the optimization is undermined by the inherent non-determinism of LLM inference. Even with deterministic decoding (zero temperature), low-level computational variations in floating-point arithmetic and parallelization introduce stochasticity to the model's outputs \citep{whitehead2017floating}. This variance creates a noisy evaluation landscape, making it difficult to reliably determine if a prompt update is genuinely effective. Second, this issue is amplified in methods that use a chain of LLM calls for feedback and updates, such as textual gradients \citep{yuksekgonul2024textgrad}. A minor variance in the initial output can be magnified as it passes through the evaluator and optimizer LLMs, a form of cascading variance \citep{dohan2022language}.

Early approaches explored the vast prompt space using search algorithms, including Monte Carlo Tree Search (e.g., PromptAgent \citep{wang2023promptagent}), genetic algorithms (e.g., GPS \citep{xu2022gps}, EvoPrompt \citep{guo2023connecting}), and discrete editing methods (e.g., GRIPS \citep{prasad2022grips}, COPLE \citep{zhan2024cople}). While innovative, these methods often struggle with the semantic complexity of language and can be sample-inefficient. More recently, iterative refinement using \textit{textual gradients} has emerged as the state-of-the-art \citep{zhang2024revolve, pryzant2023protegi, yuksekgonul2024textgrad, yu2025sipdo}. Unlike earlier search-based methods, textual-gradient approaches optimize prompts through interpretable natural-language critiques and revisions, allowing updates to follow semantic error signals rather than surface-level token edits. This improves optimization efficiency and robustness, especially on complex tasks, and has established textual-gradient-based refinement as the dominant paradigm in modern prompt optimization.

However, in this work, we 
identify and analyze two fundamental sources of instability inherent in this paradigm.
First, we find that the quality of the textual gradient is highly dependent on the correctness of the initial output. As shown in Figure \ref{fig:error} (a), generating feedback from examples that the model already handles correctly produces a low-signal and high-noise gradient, often leading to destructive edits. While this observation suggests that the solution is to optimize exclusively on the high-signal feedback from failed cases, we find it is not this straightforward, because repeated refinement on these hard examples leads to over-specialization, sacrificing the prompt's general applicability on simpler tasks.

Second, we find that optimizing exclusively on difficult cases is surprisingly unstable. While initial gradients are corrective, performance degrades sharply after just a few iterations, as shown in Figure \ref{fig:error} (b). This occurs because the optimizer attempts to solve hard examples by adding more specific constraints and multi-step procedures into the prompt. For instance, to solve a complex object-counting problem, the prompt might be amended with explicit rules like, "First, list every potential object. Second, categorize each object. Third, create a final count based only on valid categories." While this rigid algorithm is effective for the targeted hard case, it becomes overly redundant for a simple case like "count the number of apples." For simpler inputs, the verbose and complex instructions can confuse the executor model or lead to inefficient reasoning paths, thereby degrading its performance. Consequently, this leads to over-specialization and loss of generality.

\begin{figure*}[htbp]
    \centering   \includegraphics[width=0.99\textwidth,height=0.15\textheight]{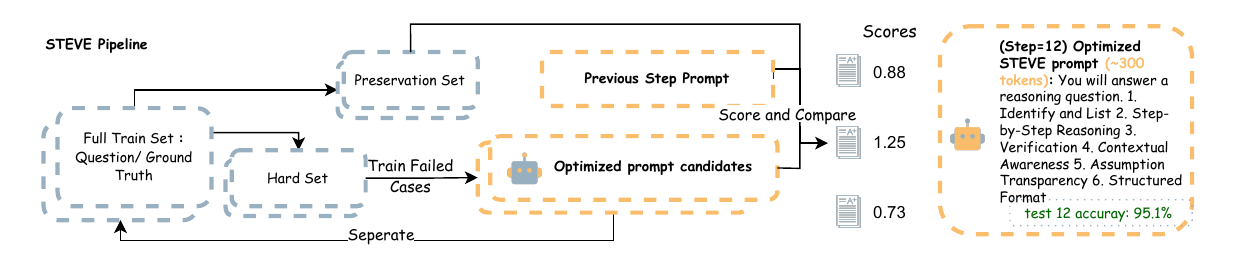}
    \vspace{-18pt}
    \caption{
    STEVE's \colorbox[HTML]{99B0C0}{Error-Driven Refinement} mechanism updates a prompt from failure cases, while \colorbox[HTML]{FABD6C}{Regularized Verification} scores multiple candidates against both the hard batch and a preservation sample. The gate selects the best non-regressing candidate, yielding a concise ($\sim$300-token) prompt with 95.1\% accuracy in this example.}
    \vspace{-0.5cm}
    \label{fig:overfit_error}
\end{figure*}

To address these two instabilities, we propose \textbf{STEVE: \underline{S}tabilizing \underline{T}extual Gradients via \underline{E}rror-Driven Refinement and Regularized \underline{Ve}rification}. Our framework, as shown in Figure \ref{fig:overfit_error}, introduces two core mechanisms. First, our \textbf{Error-Driven Refinement} strategy ensures a high-quality learning signal by exclusively generating textual gradients from model failures, filtering out the noise from correct examples. Second, to counter over-specialization, our \textbf{Regularized Verification} mechanism acts as a gate. It validates each candidate prompt against a holdout set of general examples, accepting an update only if the gain on the complex case does not compromise overall robustness. Together, these components create a stable optimization loop that effectively balances specialization and generalization.

Our contributions are threefold:
\begin{itemize}
    \item We are the first to systematically identify and analyze two primary sources of instability in iterative prompt optimization: the generation of noisy, destructive gradients from correctly-handled examples, and the rapid overfitting that occurs when optimizing exclusively on model failures.
    \item We propose \textbf{STEVE}, a simple yet effective framework that directly counteracts these instabilities in iterative prompt optimization, comprising two core mechanisms. Error-Driven Refinement ensures a high-quality learning signal by generating gradients only from model failures, while Regularized Verification acts as a novel gating mechanism that accepts a prompt update only if it preserves performance on a general holdout set. 
    \item We demonstrate through extensive experiments on complex reasoning benchmarks that our framework leads to a more stable and effective optimization process. STEVE consistently discovers robust and concise prompts that achieve state-of-the-art performance. 
    \vspace{-18pt}
\end{itemize}

\section{Method}
\label{Method}

TextGrad \citep{yuksekgonul2024textgrad} is a prominent iterative method that leverages textual gradients for prompt optimization. While we use TextGrad as our foundational optimizer (under the MIT License), the framework we introduce is data-driven and largely agnostic to the specific gradient generator, making it applicable to other iterative textual-gradient methods \citep{pryzant2023protegi, yang2023large,zhang2024revolve}. ProTeGi and TextGrad improve how critiques are generated and propagated, while REVOLVE enriches the update signal with response history. STEVE instead controls \emph{which examples produce gradients} and \emph{which candidate updates are accepted}. Neither error-only gradient generation nor the preservation-set acceptance gate is present in those methods; our ablations in Table~\ref{tab: ablation} show that both controls are necessary.
The central contribution of our work is a novel framework designed to stabilize this optimization process by directly addressing two critical failure modes we identified: the generation of noisy, often destructive gradients from correctly handled examples, and over-specialization on hard cases. In this section, we will detail the two core components of our solution: Error-Driven Refinement and Regularized Verification.
\subsection{Preliminaries: Iterative Optimization with Textual Gradients}
\label{preliminaries}
The goal of prompt optimization is to find an optimal instruction, or prompt $p^*$, that maximizes the performance of an LLM $M$ across a given task distribution $\mathcal{D}$, as expressed in \eqref{argmax}
\begin{equation}
\label{argmax}
   p^{*} = \arg\max_{p \in p_{\text{space}}} \mathbb{E}_{(x,y)\sim\mathcal{D}}[\mathcal{S}(M(p,x), y)]
\end{equation}
where $x$ is a question, $y$ is the ground-truth answer, and $\mathcal{S}$ is a task-specific evaluation metric.

To navigate this challenge, recent works \citet{zhang2024revolve, pryzant2023protegi, cui2024mapo} have proposed using an iterative refinement process guided by \textbf{textual gradients}. This approach typically employs a multi-agent framework. At each iteration $t$, the process unfolds as follows:
\begin{enumerate}[label=(\alph*)]
    \item \textbf{Forward Pass Generation:} An \textit{executor LLM} uses the current prompt, $p_t$, to process an input, $x$, and generate an output, $y = M_{executor}(p_t, x)$. 

    \item \textbf{Feedback (Gradient Calculation):} A powerful \textit{evaluator LLM} evaluates the output $y'$ and calculate the loss from $y$ and ground truth $y_{truth}$. The evaluator generates a ``textual gradient," $g_{\text{text}}$. This gradient is a natural language critique that explains the failure and provides actionable advice for improving the prompt, $g_{next} = M_{evaluator}(p_t, x \mid y, y_{truth})$
    
    \item \textbf{Update:} An \textit{optimizer LLM}, conditioned on the original prompt $p_t$ and the textual gradient $g_{\text{text}}$, synthesizes an improved prompt, $p_{t+1}$. This update step can be represented as: $p_{t+1} = M_{optimizer}(p_t, g_{\text{text}})$.
\end{enumerate}

This iterative loop allows for semantic and non-differentiable improvements to the prompt. While this powerful paradigm forms the backbone of our method, we show that this standard approach suffers from significant instability, which often prevents it from converging to a robust and high-performing solution. The following sections will analyze and address these instabilities.
\subsection{The Instability of Textual Gradient Optimization}
\label{sec:instability}

While the iterative process described in Section \ref{preliminaries} provides a powerful framework, its practical application is hindered by gradient instability.
Based on our experiment, we characterize two primary sources of this instability, which motivate our proposed control mechanisms.

\begin{figure}[htbp]
    \centering
    \includegraphics[width=\linewidth]{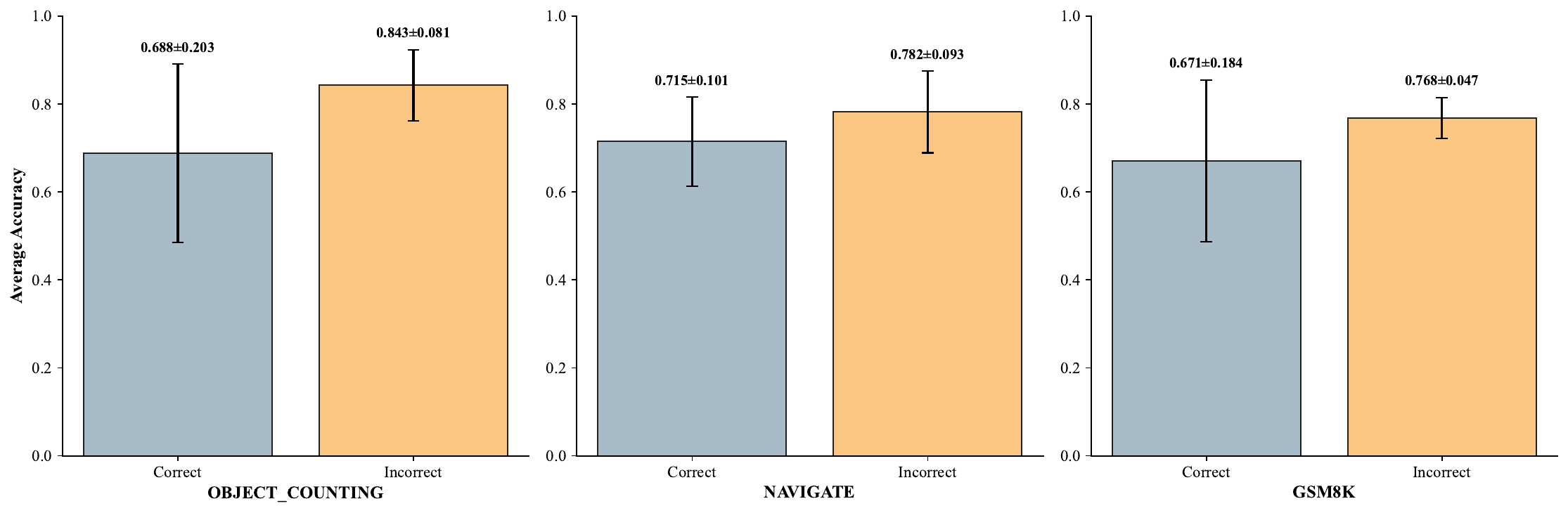}
    \caption{A comparison of single-step prompt refinement on three reasoning benchmarks Object Counting, Navigate and GSM8k \citep{suzgun2022challenging, cobbe2021training} . The bars show the final average accuracy of a prompt optimized for one round using feedback from either a batch of previously correct examples or a batch of previously incorrect examples. Across all datasets, optimizing on incorrect cases consistently yields a prompt with significantly higher accuracy, demonstrating that failures provide a superior learning signal.}
    \label{fig:triple}
    \vspace{-18pt}
\end{figure}

\paragraph{Noisy Gradients from Correct Examples.}
Our initial investigation reveals a stark contrast in the utility of feedback based on the correctness of an output. As illustrated in Figure \ref{fig:triple}, a single round of optimization using feedback from previously \textit{failed} cases yields a significant performance improvement. Conversely, using feedback from \textit{correctly-handled} cases results in a sharp drop in accuracy. This observation leads us to hypothesize that textual gradients generated from correct examples are noisy and often counter-productive. We validate this hypothesis in our ablation study (Table \ref{tab: ablation}), where a model trained exclusively on correct examples not only fails to improve but even sees its performance degrade below the initial baseline. This suggests that the evaluator LLM struggles to extract a meaningful, generalizable improvement signal from successful outputs, instead producing feedback with noisy gradients that acts as a random perturbation.
\paragraph{Over-Specialization on Hard Cases.}
Simply optimizing on failed cases, however, is not a complete solution. We observe that sustained, multi-round optimization exclusively on hard examples is also unstable, with performance degrading after an initial phase of improvement. This decay is a symptom of overfitting in the prompt space. As we show with a qualitative example in the Appendix \ref{sec:qualilative_example}, the prompt becomes progressively more verbose and convoluted over iterations, accumulating an excessive number of specific constraints and detailed steps. While these highly specific instructions may resolve the targeted hard cases, they make the prompt brittle and less effective for simpler, more general problems. The added complexity is detrimental to its overall generalization, motivating a mechanism to explicitly control this trade-off.

\subsection{Component 1: Error-Driven Refinement for High-Quality Gradients}
\label{sec:error_driven}

To address the instability caused by noisy gradients, we introduce our first mechanism: Error-Driven Refinement. Motivated by the idea of Prioritized Experience Replay\citep{schaul2015prioritized, ma2022fresher} in the field of Reinforcement Learning,  the model can focus more on transitions where its value prediction was wrong, effectively balancing the training process by focusing on hard or more informative samples. This principle that focuses on hard examples is critical for effective learning and is well-established in the machine learning literature. Seminal examples include boosting algorithms like AdaBoost\citep{freund1997decision}, which iteratively re-weights misclassified data points, and online hard example mining (OHEM)\citep{shrivastava2016training} in object detection, which explicitly trains on the most challenging examples. The core principle of this strategy is to ensure a high signal-to-noise ratio in the learning process by exclusively generating textual gradients from examples that the executor model fails to handle correctly. 

We formalize this by partitioning the training set, $\mathcal{D}_{\text{train}}$, based on the performance of the current prompt, $p_t$. Specifically, we define a ``hard case pool," $\mathcal{D}_{\text{hard}}$, as the subset of training examples where the executor model's output is deemed incorrect by the scoring function:

{\small
\begin{equation}
\label{eq:d_hard_definition}
\mathcal{D}_{\text{hard}}(p_t) = \{ (x, y) \in \mathcal{D}_{\text{train}} \mid \mathcal{S}(M_{\text{executor}}(p_t, x), y) < \tau \}
\end{equation}
}

where $\tau$ is a predefined success threshold (typically 1 for exact match tasks).

At each optimization iteration $t$, instead of sampling from the entire training distribution, our method samples an instance $(x_i, y_i)$ exclusively from this dynamically defined hard case pool, $\mathcal{D}_{\text{hard}}(p_t)$. A textual gradient, $g_{\text{text}}$, is then generated based on the model's failure on this specific instance.

By design, this error-driven strategy acts as a powerful information filter. It guarantees that every textual gradient used for an update is a high-signal, corrective piece of feedback derived from a clear failure. This eliminates the random walk behavior caused by the low-signal, high-noise gradients generated from correct examples, thereby solving the first source of instability and providing a solid foundation for targeted prompt improvement.

\begin{figure*}[t]
\centering
\begin{tcolorbox}[
    colback=lightgray,
    colframe=lightgray,
    boxsep=1pt,
    width=\textwidth,
    sharp corners,
]
    \captionof{algorithm}{The STEVE Algorithm (\underline{S}tabilizing \underline{T}extual Gradients via \underline{E}rror-Driven Refinement and Regularized \underline{Ve}rification)}
    \label{alg:steve_updated}
    \noindent\rule{\linewidth}{0.5pt} 
    \small
    \raggedright 
    
    \begin{algorithmic}[1]
        \Require Initial prompt $p_0$, training data $\mathcal{D}_{\text{train}}$, validation data $\mathcal{D}_{\text{val}}$, iterations $T$, schedule $\lambda_t$.
        \Require Preservation sample size $k$, hard-case batch size $b$, num candidates $n$.
        \Ensure Optimized prompt $p^*$.

        \State Initialize $\mathcal{D}_{\text{preserve}} \subset \{ (x, y) \in \mathcal{D}_{\text{train}} \mid \mathcal{S}(M(p_0, x), y) \geq \tau \}$.
        \State $p_t \gets p_0$

        \For{$t = 0$ to $T-1$}
            \State Define hard case pool $\mathcal{D}_{\text{hard}}(p_t) = \{ (x, y) \in \mathcal{D}_{\text{train}} \mid \mathcal{S}(M(p_t, x), y) < \tau \}$.
            \If{$|\mathcal{D}_{\text{hard}}(p_t)| < b$}
                \State \textbf{break} \Comment{Not enough errors to fix.}
            \EndIf
            \State Sample $D_{\text{batch}} = \{(x_i, y_i)\}_{i=1}^b$ from $\mathcal{D}_{\text{hard}}(p_t)$.
            \State Forward pass $\{y'_i\}_{i=1}^b \gets \{M_{\text{executor}}(p_t, x_i)\}_{i=1}^b$
            \State \Comment{\textit{Error-Driven Refinement (Section 3.3)}}
            \State $g_{\text{batch}} \gets M_{\text{evaluator}}(p_t, D_{\text{batch}}, \{y'_i\}_{i=1}^b)$.
            \State Generate $\{p_{\text{cand},j}\}_{j=1}^n$ with $M_{\text{optimizer}}(p_t,g_{\text{batch}})$.
            \State \Comment{\textit{Regularized Verification (Section 3.4)}}
            \State Sample $D^{(t)}_{\text{sample}}$ of $k=20$ examples from $\mathcal{D}_{\text{preserve}}$.
            \State $\text{best\_score} \gets 0$; $p_{t+1}\gets p_t$; $\lambda_t\gets1.5+0.1t$.
            \For{$j=1$ to $n$}
                \State $\text{Improvement} \gets \frac{1}{b}\sum_{i=1}^b (\dots)$.
                \State $\text{Regression} \gets \frac{1}{k}\sum_{(x,y)\in D^{(t)}_{\text{sample}}} (\dots)$.
                \State $\text{objective\_score} \gets \text{Improvement} - \lambda_t \max(0, \text{Regression})$.
                
                \If{$\text{objective\_score} > \text{best\_score}$}
                    \State $\text{best\_score} \gets \text{objective\_score}$
                    \State $p_{t+1} \gets p_{\text{cand},j}$
                \EndIf
            \EndFor
        \EndFor

        \State \Return $p^* = \arg \max_{p \in \{p_0, \dots, p_T\}} \mathbb{E}_{(x, y) \sim \mathcal{D}_{\text{val}}}[\mathcal{S}(M_{\text{executor}}(p, x), y)]$.
    \end{algorithmic}
\end{tcolorbox}
\vspace{-18pt}
\end{figure*}

\subsection{Component 2: Regularized Verification for Preserving Generalization}
\label{sec:regularized_verification}

While Error-Driven Refinement ensures that each textual gradient is informative, it does not prevent the optimization from over-specializing. To address this second instability, we introduce our second component: a Regularized Verification mechanism. This mechanism functions as a gate, evaluating each proposed prompt update to ensure that improvements in specialization do not come at an unacceptable cost to generalization.

Regularization is widely used to trade off task fit against robustness and generalization, including beyond continual learning in settings such as robust dataset distillation \citep{xue2025towards}. Our approach is more directly inspired by regularization-based methods in continual learning designed to combat catastrophic forgetting, most notably Elastic Weight Consolidation (EWC) \citep{kirkpatrick2017ewc}. EWC adds a quadratic penalty to the loss function to discourage modifications to network weights that are critical for performance on previously learned tasks. Analogously, our Regularized Verification mechanism treats the performance degradation on the Generalization Preservation Set as a direct penalty against forgetting. However, unlike the continuous, differentiable parameter space of neural networks where penalties can be directly integrated into a loss function, the prompt space is discrete and symbolic. Therefore, instead of modifying a loss function, we implement this regularization principle as a discrete verification step where candidate prompts are explicitly evaluated for their trade-off between specialization and generalization.

Central to this process is a fixed Generalization Preservation Pool, $\mathcal{D}_{\text{preserve}}$, constructed once from training examples that the initial prompt $p_0$ handles correctly. At iteration $t$, we draw a fresh sample $\mathcal{D}^{(t)}_{\text{sample}}$ of $k=20$ examples without replacement from this pool and reuse it to score all candidates generated at that iteration. Thus the eligible pool remains fixed, while the verification sample is resampled at every step.

{\small
\begin{equation}
\mathcal{D}_{\text{preserve}} \subset \{ (x, y) \in \mathcal{D}_{\text{train}} \mid \mathcal{S}(M_{\text{executor}}(p_0, x), y) \geq \tau \}
\end{equation}
}

The verification process is integrated into a multi-stage refinement loop. First, a batch of hard cases $\{d_h\}_1^b \subset \mathcal{D}_{\text{hard}}$ is sampled. An evaluator model synthesizes a textual gradient, $g_{\text{batch}}$, that summarizes the common failure modes across this batch. Using this gradient, an updater model generates a set of $n$ diverse candidate prompts, $\{p_{\text{cand}, i}\}_1^n$. The best candidate is then chosen based on our regularized objective. The formal update rule is:
\begin{flalign}
\label{eq:gated_update_rule}
&p_{t+1} =
\begin{cases}
p_{\text{best\_cand}} & \text{if }
    \begin{array}{@{}l@{}}
    \text{I}(p_{\text{cand}}, d_{\text{hard}}) > \\
    \quad \lambda_t \cdot \text{R}(p_{\text{cand}}, \mathcal{D}^{(t)}_{\text{sample}})
    \end{array} \\
p_t & \text{otherwise}
\end{cases}&
\end{flalign}
where I represents Improvement, R represents Regression, and $\lambda_t \geq 0$ controls the specialization--generalization trade-off. We use the linear schedule $\lambda_t=1.5+0.1t$, which makes the acceptance gate progressively more conservative; Appendix~\ref{sec:appendix_sensitivity_lambda} reports sensitivity to both its initial value and increment. By including $p_t$ among the choices, an update occurs only when a candidate improves the regularized objective. After optimization, we select the checkpoint on a held-out validation set that is disjoint from the training, preservation, and test examples; the test set is evaluated only once for the reported result. The full process is detailed in Algorithm~\ref{alg:steve_updated}.

\section{Experiment}
\label{sec:experiments}

We evaluate whether STEVE outperforms established prompt-optimization methods across diverse reasoning tasks, isolate the contribution of each component, and study sensitivity to the evaluator/optimizer and regularization schedule. Additional results and implementation details appear in the Appendix. We will release the source code and evaluation data with the paper.

\begin{table*}[t!]
\centering
\renewcommand{\arraystretch}{1.2} 
\resizebox{\textwidth}{!}{%
\begin{tabular}{@{}lll c @{\hskip 1em}c @{\hskip 2em} c c c@{}}
\toprule
& & & \multicolumn{2}{c}{\textbf{Baselines}} & \multicolumn{3}{c}{\textbf{Iterative Optimization Methods}} \\
\cmidrule(lr){4-5} \cmidrule(l){6-8}
\textbf{Category} & \textbf{Task} & \textbf{Evaluator/Optimizer Model} & Zero-shot & Three-shot & TextGrad & REVOLVE & STEVE (Ours) \\
\midrule
\multirow{6}{*}{\rotatebox[origin=c]{90}{\parbox[c]{1.5cm}{\centering Math}}}
& \multirow{3}{*}{\makecell{GSM8k \\ \small\citep{cobbe2021training}}} & \texttt{gpt-4o} & 76.3 & 73.1 & 82.5\,\gain{+6.2} & \underline{82.8}\,\gain{+6.5} & \textbf{86.2}\,\gain{+9.9} \\
& & \texttt{gemini-2.5-flash} & 73.8 & 74.2 & 81.1\,\gain{+7.3} & \textbf{83.9}\,\gain{+10.1} & \underline{82.9}\,\gain{+9.1} \\
& & \texttt{gpt-5} & 73.2 & 77.5 & 78.6\,\gain{+5.4} & \underline{81.7}\,\gain{+8.5} & \textbf{84.3}\,\gain{+11.1} \\
\cmidrule(l){2-8}
& \multirow{3}{*}{\makecell{MultiArith \\ \small\citep{roy2016solving}}} & \texttt{gpt-4o} & 84.5 & 84.1 & 88.6\,\gain{+4.1} & \underline{98.2}\,\gain{+13.7} & \textbf{98.9}\,\gain{+14.4} \\
& & \texttt{gemini-2.5-flash} & 82.9 & 84.3 & 98.1\,\gain{+15.2} & \underline{98.4}\,\gain{+15.5} & \textbf{98.7}\,\gain{+15.8} \\
& & \texttt{gpt-5} & 84.0 & 84.6 & \textbf{100.0}\,\gain{+16.0} & \textbf{100.0}\,\gain{+16.0} & \textbf{100.0}\,\gain{+16.0} \\
\midrule
\multirow{6}{*}{\rotatebox[origin=c]{90}{\parbox[c]{2.5cm}{\centering Commonsense}}}
& \multirow{3}{*}{\makecell{StrategyQA \\ \small\citep{geva2021did}}} & \texttt{gpt-4o} & 88.7 & \underline{91.3} & 90.1\,\gain{+1.4} & 90.5\,\gain{+1.8} & \textbf{93.2}\,\gain{+4.5} \\
& & \texttt{gemini-2.5-flash} & 85.1 & \underline{90.6} & 88.8\,\gain{+3.7} & 89.4\,\gain{+4.3} & \textbf{91.9}\,\gain{+6.8} \\
& & \texttt{gpt-5} & 88.2 & \underline{94.5} & 91.7\,\gain{+3.5} & 93.3\,\gain{+5.1} & \textbf{95.8}\,\gain{+7.6} \\
\cmidrule(l){2-8}
& \multirow{3}{*}{\makecell{Navigate \\ \small\citep{suzgun2022challenging}}} & \texttt{gpt-4o} & 60.1 & 77.8 & 88.2\,\gain{+28.1} & \underline{94.1}\,\gain{+34.0} & \textbf{95.6}\,\gain{+35.5} \\
& & \texttt{gemini-2.5-flash} & 68.7 & 83.6 & 66.9\,\loss{-1.8} & \underline{90.4}\,\gain{+21.7} & \textbf{96.2}\,\gain{+27.5} \\
& & \texttt{gpt-5} & 62.7 & 83.1 & 78.6\,\gain{+15.9} & \textbf{86.1}\,\gain{+23.4} & \underline{83.9}\,\gain{+21.2} \\
\midrule
\multirow{12}{*}{\rotatebox[origin=c]{90}{\parbox[c]{2.5cm}{\centering Symbolic}}}
& \multirow{3}{*}{\makecell{Object Counting \\ \small\citep{suzgun2022challenging}}} & \texttt{gpt-4o} & 77.9 & 82.2 & 87.1\,\gain{+9.2} & \underline{90.3}\,\gain{+12.4} & \textbf{95.7}\,\gain{+17.8} \\
& & \texttt{gemini-2.5-flash} & 75.4 & \underline{81.8} & 72.5\,\loss{-2.9} & 81.1\,\gain{+5.7} & \textbf{83.6}\,\gain{+8.2} \\
& & \texttt{gpt-5} & 79.0 & \underline{87.5} & 80.6\,\gain{+1.6} & 82.4\,\gain{+3.4} & \textbf{91.2}\,\gain{+12.2} \\
\cmidrule(l){2-8}
& \multirow{3}{*}{\makecell{Penguins in a Table \\ \small\citep{suzgun2022challenging}}} & \texttt{gpt-4o} & 80.8 & 83.3 & 93.1\,\gain{+12.3} & \underline{96.0}\,\gain{+15.2} & \textbf{96.5}\,\gain{+15.7} \\
& & \texttt{gemini-2.5-flash} & 66.2 & 90.7 & \underline{90.9}\,\gain{+24.7} & 90.2\,\gain{+24.0} & \textbf{96.4}\,\gain{+30.2} \\
& & \texttt{gpt-5} & 63.5 & 90.1 & \underline{93.6}\,\gain{+30.1} & 86.8\,\gain{+23.3} & \textbf{96.7}\,\gain{+33.2} \\
\cmidrule(l){2-8}
& \multirow{3}{*}{\makecell{Geometric Shapes \\ \small\citep{suzgun2022challenging}}} & \texttt{gpt-4o} & 39.4 & 31.8 & 36.7\,\loss{-2.7} & \underline{55.2}\,\gain{+15.8} & \textbf{62.9}\,\gain{+23.5} \\
& & \texttt{gemini-2.5-flash} & 36.6 & 36.1 & 48.3\,\gain{+11.7} & \underline{65.5}\,\gain{+28.9} & \textbf{66.0}\,\gain{+29.4} \\
& & \texttt{gpt-5} & 41.1 & 32.7 & 33.9\,\loss{-7.2} & \underline{42.4}\,\gain{+1.3} & \textbf{44.2}\,\gain{+3.1} \\
\cmidrule(l){2-8}
& \multirow{3}{*}{\makecell{Date Understanding \\ \small\citep{suzgun2022challenging}}} & \texttt{gpt-4o} & 67.3 & 73.9 & 75.1\,\gain{+7.8} & \underline{76.2}\,\gain{+8.9} & \textbf{76.6}\,\gain{+9.3} \\
& & \texttt{gemini-2.5-flash} & 70.8 & 68.2 & 74.4\,\gain{+3.6} & \underline{74.9}\,\gain{+4.1} & \textbf{80.5}\,\gain{+9.7} \\
& & \texttt{gpt-5} & 70.1 & 72.7 & 74.3\,\gain{+4.2} & \underline{77.8}\,\gain{+7.7} & \textbf{84.0}\,\gain{+13.9} \\
\midrule
\multirow{6}{*}{\rotatebox[origin=c]{90}{\parbox[c]{2cm}{\centering Expert}}}
& \multirow{3}{*}{\makecell{College Physics \\ \small\citep{hendrycks2020measuring}}} & \texttt{gpt-4o} & 57.6 & 52.1 & 61.4\,\gain{+3.8} & \underline{66.9}\,\gain{+9.3} & \textbf{71.2}\,\gain{+13.6} \\
& & \texttt{gemini-2.5-flash} & 61.0 & 52.8 & 57.3\,\loss{-3.7} & \underline{61.5}\,\gain{+0.5} & \textbf{66.7}\,\gain{+5.7} \\
& & \texttt{gpt-5} & 52.3 & 57.9 & 57.5\,\gain{+5.2} & \underline{61.8}\,\gain{+9.5} & \textbf{66.1}\,\gain{+13.8} \\
\cmidrule(l){2-8}
& \multirow{3}{*}{\makecell{Machine Learning \\ \small\citep{hendrycks2020measuring}}} & \texttt{gpt-4o} & 34.2 & 39.8 & 60.1\,\gain{+25.9} & \underline{60.3}\,\gain{+26.2} & \textbf{60.7}\,\gain{+26.5} \\
& & \texttt{gemini-2.5-flash} & 43.5 & 34.6 & 52.9\,\gain{+9.4} & \textbf{60.4}\,\gain{+16.9} & \underline{56.2}\,\gain{+12.7} \\
& & \texttt{gpt-5} & 47.7 & 39.1 & \underline{56.6}\,\gain{+8.9} & 43.3\,\loss{-4.4} & \textbf{60.9}\,\gain{+13.2} \\
\bottomrule
\end{tabular}%
}
\caption{Main results on \textbf{Mathematical, Commonsense, Symbolic/Procedural,} and \textbf{Expert-Level Knowledge Reasoning} benchmarks, broken down by evaluator/optimizer model. The executor is \texttt{gpt-3.5-turbo-0125} in every row. All iterative methods are optimized for 12 steps. We report accuracy with the absolute change ($\uparrow\downarrow$) from Zero-shot CoT. Bold and underlining denote the best and second-best result in each row.}
\label{tab:main_results}
\vspace{-0.4cm}
\end{table*}

\subsection{Datasets and Tasks}
\label{sec:datasets}

To ensure a thorough assessment of our method's generalization capabilities, we evaluate it on 10 challenging benchmarks spanning four distinct reasoning domains. For mathematical reasoning, we use \textbf{GSM8k} \citep{cobbe2021training} and \textbf{MultiArith} \citep{roy2016solving}, which test multi-step numerical problem-solving. For complex commonsense reasoning, we assess multi-hop logical deduction using \textbf{StrategyQA} \citep{geva2021did} and the \textbf{Navigate} task from Big-Bench Hard (BBH) \citep{suzgun2022challenging}. To evaluate precise procedural execution in symbolic and procedural reasoning, we use four tasks from BBH: \textbf{Object Counting}, \textbf{Penguins in a Table}, \textbf{Geometric Shapes}, and \textbf{Date Understanding}. Finally, to assess expert-level knowledge reasoning on domain-specific topics, we use the \textbf{College Physics} and \textbf{Machine Learning} subsets from the Massive Multitask Language Understanding (MMLU) benchmark 
\citep{hendrycks2020measuring}.
\subsection{Experimental Setup}
\label{sec:setup}

\paragraph{Baselines.}
Our original evaluation compares STEVE with four baselines. \textbf{Zero-shot CoT} \citep{kojima2022large} supplies the initial prompt $P_0$ for every iterative method, while \textbf{Three-shot CoT} \citep{wei2022chain} is a manually constructed few-shot baseline. \textbf{TextGrad} \citep{yuksekgonul2024textgrad} uses natural-language feedback as a gradient to refine prompts, and \textbf{REVOLVE} \citep{zhang2024revolve} conditions updates on the historical evolution of responses. Our modernized evaluation additionally includes \textbf{ProTeGi} \citep{pryzant2023protegi}, which combines textual gradients with beam search and bandit selection. Appendix~\ref{sec:appendix_evoprompt_comparison} reports a non-gradient comparison with EvoPrompt.

\paragraph{Implementation Details.}
For the original 10-benchmark evaluation in Table~\ref{tab:main_results}, \texttt{gpt-3.5-turbo-0125} is the \emph{sole executor}; \texttt{gpt-4o}, \texttt{gemini-2.5-flash}, and \texttt{gpt-5} serve only as the evaluator/optimizer alternatives shown in the table. All iterative methods run for $T=12$ steps with hard-case batch size $b=4$, preservation sample size $k=20$, and three candidates per step. We use temperature 0.0 and report mean accuracy over three independent runs. Checkpoints are selected using held-out validation data; test examples are not used for prompt selection. Appendix~\ref{sec:appendix_implementation} gives further details.

\subsection{Results}
\label{sec:results}

We present our primary findings in Table \ref{tab:main_results}, which compares the final accuracy of STEVE against all baselines across our 10 benchmark datasets and three evaluator/optimizer models. The results demonstrate that STEVE outperforms both static baselines (Zero-shot and Three-shot CoT) and state-of-the-art iterative optimization methods. Averaged across all 30 settings, STEVE achieves an absolute improvement of over 15\% compared to the initial Zero-shot CoT prompt and outperforms the strongest iterative baseline, REVOLVE, by an average of 3.5\%.

STEVE shows particularly strong performance on tasks requiring complex procedural or symbolic reasoning. For instance, on Navigate and Penguins in a Table with the \texttt{gemini-2.5-flash} evaluator/optimizer, STEVE achieves gains of over 27\% and 30\%, respectively. We hypothesize that these tasks involve discovering non-obvious, robust strategies that are easily missed by unstable optimizers. STEVE's verification mechanism allows it to safely explore and lock in complex heuristics that generalize well, whereas other methods may discard them or overfit to a brittle solution. This finding further demonstrates the robustness and effectiveness of our STEVE method.

While broadly successful, the margin of improvement varies. On tasks such as MultiArith, where the initial prompt is already quite effective, the gains are more modest as there is less room for optimization. 
\newcommand{\methodheader}[1]{\multicolumn{3}{c}{#1}}

\paragraph{Modernized models, benchmarks, and baselines.}
To test whether the gains persist beyond the smaller executor and benchmark subsets used above, we reran the comparison with \texttt{gpt-5.4-mini} as executor and \texttt{gpt-5.4} as evaluator/optimizer. Table~\ref{tab:modern_results} adds ProTeGi and covers two tasks from BIG-Bench Extra Hard (BBEH) \citep{kazemi2025bbeh}, the full 1,009-example GSM8K-Platinum evaluation \citep{vendrow2025platinum}, and 800 DS-1000 code-generation problems \citep{lai2023ds1000}. STEVE performs best on all four tasks. Its largest margins over TextGrad are on the harder symbolic tasks: +8.0 points on Object Counting and +10.0 on Geometric Shapes. Gains are smaller on the nearly saturated GSM8K-Platinum benchmark (+0.1) and DS-1000 (+0.5).

\begin{table*}[t]
\centering
\small
\setlength{\tabcolsep}{3pt}
\begin{tabular}{llccrrrrr}
\toprule
\textbf{Category} & \textbf{Task} & \textbf{Executor} & \textbf{Optimizer} & \textbf{Zero-shot} & \textbf{Three-shot} & \textbf{ProTeGi} & \textbf{TextGrad} & \textbf{STEVE} \\
\midrule
Symbolic & BBEH Object Counting & \texttt{gpt-5.4-mini} & \texttt{gpt-5.4} & 18.0 & 14.0 & 20.0 & 44.0 & \textbf{52.0} \\
Symbolic & BBEH Geometric Shapes & \texttt{gpt-5.4-mini} & \texttt{gpt-5.4} & 24.0 & 14.0 & 32.0 & 34.0 & \textbf{44.0} \\
Math & GSM8K-Platinum & \texttt{gpt-5.4-mini} & \texttt{gpt-5.4} & 98.0 & 98.1 & 98.2 & 98.2 & \textbf{98.3} \\
Code & DS-1000 & \texttt{gpt-5.4-mini} & \texttt{gpt-5.4} & 27.5 & 27.5 & 32.1 & 32.0 & \textbf{32.5} \\
\bottomrule
\end{tabular}
\caption{Accuracy (\%) in the modernized evaluation. All methods use the same executor/evaluator configuration. GSM8K-Platinum and DS-1000 results use 1,009 and 800 test examples, respectively.}
\label{tab:modern_results}
\end{table*}

\subsection{Ablation Studies}
\label{sec:ablations}

To isolate and validate the contributions of our framework's core components, we conduct a series of ablation studies on a representative subset of datasets: GSM8k, StrategyQA, and Object Counting. The results, summarized in Table \ref{tab: ablation}, confirm our design choices.

\begin{table}[h!]
\centering
\resizebox{\columnwidth}{!}{%
\begin{tabular}{lccc}
\toprule
\textbf{Model Variant} & \textbf{GSM8k} & \textbf{StrategyQA} & \textbf{Object Counting} \\
\midrule
Initial CoT Prompt & 76.3 & 88.7 & 77.9 \\
\midrule
\textit{Regularized Verification} \\
\quad w/o Verification & 84.1 & 91.5 & 92.2 \\
\midrule
\textit{Error-Driven Refinement} \\
\quad w/ Full Dataset & 82.5 & 90.1 & 87.1 \\
\quad w/ Correct-Only & 76.5 & 86.8 & 76.1 \\
\midrule
STEVE (Ours) & \textbf{86.2} & \textbf{93.2} & \textbf{95.7} \\
\bottomrule
\end{tabular}}
\caption{Ablation studies on the core components of STEVE. We report performance on a representative subset of datasets. The results demonstrate the importance of both the regularized verification gate and the error-driven learning signal. The best performance for each dataset is highlighted in \textbf{bold}.}
\label{tab: ablation}
\vspace{-14pt}
\end{table}

\paragraph{Contribution of Regularized Verification.}
First, to demonstrate the necessity of our verification gate, we evaluate STEVE w/o Verification. In this variant, we remove the Verification module, meaning the candidate prompts are trained by the hard-case batch without a regression check. This variant underperforms the full STEVE model, confirming that without explicit regularization to safeguard general capabilities, the prompt quickly overfits to the hard cases it is trained on.

\paragraph{Analysis of the Error-Driven Signal.}
Second, we validate our core hypothesis that the learning signal must be error-driven. We test two variants: (1) STEVE w/ Full Dataset, which trains on the complete, unfiltered training set containing both correct and incorrect examples, and (2) STEVE w/ Correct-Only, a control experiment that trains exclusively on examples the model already handles correctly. The first variant suffers from the noisy gradients of correct examples and performs poorly. The second variant consistently degrades the prompt's performance, often below the initial baseline. These results provide strong evidence that a high-quality, corrective signal derived exclusively from errors is essential for stable and effective optimization.

\section{Conclusion}
In this work, we address the critical instability of textual gradient-based prompt optimization by reframing it as a dual-selection problem. To solve the failure modes of noisy feedback and over-specialization, we introduced STEVE, a framework that carefully picks its updates at two key stages. First, our Error-Driven Refinement mechanism ensures a high-quality learning signal by selectively score feedback generated exclusively from previously failed cases, filtering out the noise from correct examples. Second, our Regularized Verification step addresses overfitting by selecting the best candidate prompt. Our experiments demonstrated that this principled approach of carefully picking both the learning signal and the final update allows STEVE to consistently discover more robust and effective prompts than standard methods.

\section*{Limitations}
Our approach is inherently model-dependent, as it uses a loop where an evaluator produces a textual gradient and an optimizer converts that feedback into prompt updates. As a result, the optimization quality and stability can vary with the choice of model families and their specific API versions used in the experiments. Small changes in such proprietary models may lead to different updated prompts, limiting portability and making outcomes sensitive to implementation details. Moreover, the method incurs substantial computational cost. To improve stability, it generates multiple diverse candidates per iteration and performs a regularized verification by evaluating each candidate on a preservation set, introducing an overhead of additional executor calls per iteration compared to baseline approaches.

\section*{Ethical Considerations}
This work evaluates prompt optimization on publicly available reasoning and code-generation benchmarks and does not collect personal data or involve human subjects. Nevertheless, STEVE relies on both proprietary and open-source large language models, and optimized prompts and model outputs may inherit biases, factual errors, or unsafe behaviors from the underlying models. Performance on the benchmarks studied here should therefore not be interpreted as a guarantee of reliability in high-stakes or safety-critical settings. The iterative optimization procedure also requires repeated model calls for candidate generation and verification, resulting in additional financial and environmental costs; we report model versions, token usage, API-call counts, and estimated cost where available to improve transparency. Practitioners should follow applicable model licenses and terms of use, evaluate optimized prompts on representative preservation sets, and retain appropriate human oversight before deployment.

\bibliography{main}

\begin{thebibliography}{51}
\providecommand{\natexlab}[1]{#1}

\bibitem[{Cobbe et~al.(2021)Cobbe, Kosaraju, Bavarian, Chen, Jun, Kaiser,
  Plappert, Tworek, Hilton, Nakano et~al.}]{cobbe2021training}
Karl Cobbe, Vineet Kosaraju, Mohammad Bavarian, Mark Chen, Heewoo Jun, Lukasz
  Kaiser, Matthias Plappert, Jerry Tworek, Jacob Hilton, Reiichiro Nakano,
  et~al. 2021.
\newblock Training verifiers to solve math word problems.
\newblock \emph{arXiv preprint arXiv:2110.14168}.

\bibitem[{Cui et~al.(2024)Cui, Nandyalam, Rufail, Cheung, Lei, Zhu, and
  O'Brien}]{cui2024mapo}
Anthony Cui, Pranav Nandyalam, Andrew Rufail, Ethan Cheung, Aiden Lei, Kevin
  Zhu, and Sean O'Brien. 2024.
\newblock Introducing mapo: Momentum-aided gradient descent prompt
  optimization.
\newblock \emph{arXiv preprint arXiv:2410.19499}.

\bibitem[{Deng et~al.(2022)Deng, Wang, Hsieh, Wang, Guo, Shu, Song, Xing, and
  Hu}]{deng2022rlprompt}
Mingkai Deng, Jianyu Wang, Cheng-Ping Hsieh, Yihan Wang, Han Guo, Tianmin Shu,
  Meng Song, Eric~P Xing, and Zhiting Hu. 2022.
\newblock Rlprompt: Optimizing discrete text prompts with reinforcement
  learning.
\newblock \emph{arXiv preprint arXiv:2205.12548}.

\bibitem[{Do et~al.(2024)Do, Zhao, Brown, Xie, Zhao, Chen, Kawaguchi, Shieh,
  and He}]{adv-ICL}
Xuan~Long Do, Yiran Zhao, Hannah Brown, Yuxi Xie, James~Xu Zhao, Nancy Chen,
  Kenji Kawaguchi, Michael Shieh, and Junxian He. 2024.
\newblock Prompt optimization via adversarial in-context learning.
\newblock In \emph{Proceedings of the 62nd Annual Meeting of the Association
  for Computational Linguistics (Volume 1: Long Papers)}, pages 7308--7327.

\bibitem[{Dohan et~al.(2022)Dohan, Xu, Lewkowycz, Austin, Bieber, Lopes, Wu,
  Michalewski, Saurous, Sohl-Dickstein et~al.}]{dohan2022language}
David Dohan, Winnie Xu, Aitor Lewkowycz, Jacob Austin, David Bieber,
  Raphael~Gontijo Lopes, Yuhuai Wu, Henryk Michalewski, Rif~A Saurous, Jascha
  Sohl-Dickstein, et~al. 2022.
\newblock Language model cascades.
\newblock \emph{arXiv preprint arXiv:2207.10342}.

\bibitem[{Freund and Schapire(1997)}]{freund1997decision}
Yoav Freund and Robert~E Schapire. 1997.
\newblock A decision-theoretic generalization of on-line learning and an
  application to boosting.
\newblock \emph{Journal of computer and system sciences}, 55(1):119--139.

\bibitem[{Geva et~al.(2021)Geva, Khashabi, Segal, Khot, Roth, and
  Berant}]{geva2021did}
Mor Geva, Daniel Khashabi, Elad Segal, Tushar Khot, Dan Roth, and Jonathan
  Berant. 2021.
\newblock Did aristotle use a laptop? a question answering benchmark with
  implicit reasoning strategies.
\newblock \emph{Transactions of the Association for Computational Linguistics},
  9:346--361.

\bibitem[{Guo et~al.(2024)Guo, Wang, Guo, Li, Song, Tan, Liu, Bian, and
  Yang}]{guo2023connecting}
Qingyan Guo, Rui Wang, Junliang Guo, Bei Li, Kaitao Song, Xu~Tan, Guoqing Liu,
  Jiang Bian, and Yujiu Yang. 2024.
\newblock \href {https://openreview.net/forum?id=ZG3RaNIsO8} {Connecting large
  language models with evolutionary algorithms yields powerful prompt
  optimizers}.
\newblock In \emph{The Twelfth International Conference on Learning
  Representations}.

\bibitem[{Hendrycks et~al.(2020)Hendrycks, Burns, Basart, Zou, Mazeika, Song,
  and Steinhardt}]{hendrycks2020measuring}
Dan Hendrycks, Collin Burns, Steven Basart, Andy Zou, Mantas Mazeika, Dawn
  Song, and Jacob Steinhardt. 2020.
\newblock Measuring massive multitask language understanding.
\newblock \emph{arXiv preprint arXiv:2009.03300}.

\bibitem[{Hu et~al.(2022)Hu, Shen, Wallis, Allen-Zhu, Li, Wang, Wang, Chen
  et~al.}]{hu2022lora}
Edward~J Hu, Yelong Shen, Phillip Wallis, Zeyuan Allen-Zhu, Yuanzhi Li, Shean
  Wang, Lu~Wang, Weizhu Chen, et~al. 2022.
\newblock Lora: Low-rank adaptation of large language models.
\newblock \emph{ICLR}, 1(2):3.

\bibitem[{Jain and Jindal(2025)}]{11106067}
Aditi~M Jain and Mayank Jindal. 2025.
\newblock Systematic survey of various prompt optimization methods and their
  classifications.
\newblock In \emph{2025 11th International Conference on Computing and
  Artificial Intelligence (ICCAI)}, pages 524--536.

\bibitem[{Kazemi et~al.(2025)Kazemi, Fatemi, Bansal, Palowitch, Anastasiou,
  Mehta, Jain, Aglietti, Jindal, Chen, Dikkala, Tyen, Liu, Shalit, Chiappa,
  Olszewska, Tay, Tran, Le, and Firat}]{kazemi2025bbeh}
Mehran Kazemi, Bahare Fatemi, Hritik Bansal, John Palowitch, Chrysovalantis
  Anastasiou, Sanket~Vaibhav Mehta, Lalit~K Jain, Virginia Aglietti, Disha
  Jindal, Peter Chen, Nishanth Dikkala, Gladys Tyen, Xin Liu, Uri Shalit,
  Silvia Chiappa, Kate Olszewska, Yi~Tay, Vinh~Q. Tran, Quoc~V Le, and Orhan
  Firat. 2025.
\newblock \href {https://doi.org/10.18653/v1/2025.acl-long.1285} {{BIG}-bench
  extra hard}.
\newblock In \emph{Proceedings of the 63rd Annual Meeting of the Association
  for Computational Linguistics (Volume 1: Long Papers)}, pages 26473--26501,
  Vienna, Austria. Association for Computational Linguistics.

\bibitem[{Kirkpatrick et~al.(2017)Kirkpatrick, Pascanu, Rabinowitz, Veness,
  Desjardins, Rusu, Milan, Quan, Ramalho, Grabska-Barwinska
  et~al.}]{kirkpatrick2017ewc}
James Kirkpatrick, Razvan Pascanu, Neil Rabinowitz, Joel Veness, Guillaume
  Desjardins, Andrei~A Rusu, Kieran Milan, John Quan, Tiago Ramalho, Agnieszka
  Grabska-Barwinska, et~al. 2017.
\newblock Overcoming catastrophic forgetting in neural networks.
\newblock \emph{Proceedings of the national academy of sciences},
  114(13):3521--3526.

\bibitem[{Kojima et~al.(2022)Kojima, Gu, Reid, Matsuo, and
  Iwasawa}]{kojima2022large}
Takeshi Kojima, Shixiang~Shane Gu, Machel Reid, Yutaka Matsuo, and Yusuke
  Iwasawa. 2022.
\newblock Large language models are zero-shot reasoners.
\newblock \emph{Advances in neural information processing systems},
  35:22199--22213.

\bibitem[{Lai et~al.(2023)Lai, Li, Wang, Zhang, Zhong, Zettlemoyer, Yih, Fried,
  Wang, and Yu}]{lai2023ds1000}
Yuhang Lai, Chengxi Li, Yiming Wang, Tianyi Zhang, Ruiqi Zhong, Luke
  Zettlemoyer, Wen-Tau Yih, Daniel Fried, Sida Wang, and Tao Yu. 2023.
\newblock \href {https://proceedings.mlr.press/v202/lai23b.html} {{DS}-1000: A
  natural and reliable benchmark for data science code generation}.
\newblock In \emph{Proceedings of the 40th International Conference on Machine
  Learning}, volume 202 of \emph{Proceedings of Machine Learning Research},
  pages 18319--18345. PMLR.

\bibitem[{Lester et~al.(2021)Lester, Al-Rfou, and Constant}]{lestersoftprompt}
Brian Lester, Rami Al-Rfou, and Noah Constant. 2021.
\newblock The power of scale for parameter-efficient prompt tuning.
\newblock \emph{arXiv preprint arXiv:2104.08691}.

\bibitem[{Li et~al.(2024)Li, Wang, Chen, Jiang, Ding, and
  Okumura}]{li2024survey}
Dongyuan Li, Zhen Wang, Yankai Chen, Renhe Jiang, Weiping Ding, and Manabu
  Okumura. 2024.
\newblock A survey on deep active learning: Recent advances and new frontiers.
\newblock \emph{IEEE Transactions on Neural Networks and Learning Systems},
  36(4):5879--5899.

\bibitem[{Liu et~al.(2026)Liu, Wang, Li, Zhu, Shen, Wang, Abbasi, Zhang, and
  Ji}]{liu2026augmenting}
Jiateng Liu, Rushi Wang, Bingxuan Li, Kunlun Zhu, Yifan Shen, Qingyun Wang,
  Ahmed Abbasi, Denghui Zhang, and Heng Ji. 2026.
\newblock \href {https://arxiv.org/abs/2605.02729} {Augmenting interface
  usability heuristics for reliable computer-use agents}.
\newblock \emph{arXiv preprint arXiv:2605.02729}.

\bibitem[{Liu et~al.(2024)Liu, Zheng, Du, Ding, Qian, Yang, and
  Tang}]{liu2024gpt}
Xiao Liu, Yanan Zheng, Zhengxiao Du, Ming Ding, Yujie Qian, Zhilin Yang, and
  Jie Tang. 2024.
\newblock Gpt understands, too.
\newblock \emph{AI Open}, 5:208--215.

\bibitem[{Ma et~al.(2022)Ma, Ning, Zhang, and Liu}]{ma2022fresher}
Jue Ma, Dejun Ning, Chengyi Zhang, and Shipeng Liu. 2022.
\newblock Fresher experience plays a more important role in prioritized
  experience replay.
\newblock \emph{Applied sciences}, 12(23):12489.

\bibitem[{Nguyen et~al.(2021)Nguyen, Low, and Jaillet}]{nguyen2021information}
Quoc~Phong Nguyen, Bryan Kian~Hsiang Low, and Patrick Jaillet. 2021.
\newblock An information-theoretic framework for unifying active learning
  problems.
\newblock In \emph{Proceedings of the AAAI Conference on Artificial
  Intelligence}, volume~35, pages 9126--9134.

\bibitem[{Park et~al.(2023)Park, O'Brien, Cai, Morris, Liang, and
  Bernstein}]{park2023generative}
Joon~Sung Park, Joseph O'Brien, Carrie~Jun Cai, Meredith~Ringel Morris, Percy
  Liang, and Michael~S Bernstein. 2023.
\newblock Generative agents: Interactive simulacra of human behavior.
\newblock In \emph{Proceedings of the 36th annual acm symposium on user
  interface software and technology}, pages 1--22.

\bibitem[{Prasad et~al.(2022)Prasad, Hase, Zhou, and Bansal}]{prasad2022grips}
Archiki Prasad, Peter Hase, Xiang Zhou, and Mohit Bansal. 2022.
\newblock Grips: Gradient-free, edit-based instruction search for prompting
  large language models.
\newblock \emph{arXiv preprint arXiv:2203.07281}.

\bibitem[{Pryzant et~al.(2023)Pryzant, Iter, Li, Lee, Zhu, and
  Zeng}]{pryzant2023protegi}
Reid Pryzant, Dan Iter, Jerry Li, Yin Lee, Chenguang Zhu, and Michael Zeng.
  2023.
\newblock \href {https://doi.org/10.18653/v1/2023.emnlp-main.494} {Automatic
  prompt optimization with ``gradient descent'' and beam search}.
\newblock In \emph{Proceedings of the 2023 Conference on Empirical Methods in
  Natural Language Processing}, pages 7957--7968, Singapore. Association for
  Computational Linguistics.

\bibitem[{Roy and Roth(2016)}]{roy2016solving}
Subhro Roy and Dan Roth. 2016.
\newblock Solving general arithmetic word problems.
\newblock \emph{arXiv preprint arXiv:1608.01413}.

\bibitem[{Schaul et~al.(2015)Schaul, Quan, Antonoglou, and
  Silver}]{schaul2015prioritized}
Tom Schaul, John Quan, Ioannis Antonoglou, and David Silver. 2015.
\newblock Prioritized experience replay.
\newblock \emph{arXiv preprint arXiv:1511.05952}.

\bibitem[{Schick et~al.(2023)Schick, Dwivedi-Yu, Dess{\`\i}, Raileanu, Lomeli,
  Hambro, Zettlemoyer, Cancedda, and Scialom}]{schick2023toolformer}
Timo Schick, Jane Dwivedi-Yu, Roberto Dess{\`\i}, Roberta Raileanu, Maria
  Lomeli, Eric Hambro, Luke Zettlemoyer, Nicola Cancedda, and Thomas Scialom.
  2023.
\newblock Toolformer: Language models can teach themselves to use tools.
\newblock \emph{Advances in Neural Information Processing Systems},
  36:68539--68551.

\bibitem[{Settles(2009)}]{settles2009active}
Burr Settles. 2009.
\newblock Active learning literature survey.
\newblock Technical report, University of Wisconsin--Madison Department of
  Computer Sciences.

\bibitem[{Shen et~al.(2025)Shen, Liu, Zhu, Cao, Zhang, He, Ye, Rehg, and
  Lourentzou}]{shen2025fine}
Yifan Shen, Yuanzhe Liu, Jingyuan Zhu, Xu~Cao, Xiaofeng Zhang, Yixiao He,
  Wenming Ye, James~M. Rehg, and Ismini Lourentzou. 2025.
\newblock \href {https://doi.org/10.52202/085713-0606} {Fine-grained preference
  optimization improves spatial reasoning in {VLM}s}.
\newblock In \emph{Advances in Neural Information Processing Systems},
  volume~38.

\bibitem[{Shrivastava et~al.(2016)Shrivastava, Gupta, and
  Girshick}]{shrivastava2016training}
Abhinav Shrivastava, Abhinav Gupta, and Ross Girshick. 2016.
\newblock Training region-based object detectors with online hard example
  mining.
\newblock In \emph{Proceedings of the IEEE conference on computer vision and
  pattern recognition}, pages 761--769.

\bibitem[{Sinha et~al.(2024)Sinha, Cui, Das, and Zhang}]{sinha2024survival}
Ankita Sinha, Wendi Cui, Kamalika Das, and Jiaxin Zhang. 2024.
\newblock Survival of the safest: Towards secure prompt optimization through
  interleaved multi-objective evolution.
\newblock \emph{arXiv preprint arXiv:2410.09652}.

\bibitem[{Suzgun et~al.(2022)Suzgun, Scales, Sch{\"a}rli, Gehrmann, Tay, Chung,
  Chowdhery, Le, Chi, Zhou et~al.}]{suzgun2022challenging}
Mirac Suzgun, Nathan Scales, Nathanael Sch{\"a}rli, Sebastian Gehrmann, Yi~Tay,
  Hyung~Won Chung, Aakanksha Chowdhery, Quoc~V Le, Ed~H Chi, Denny Zhou, et~al.
  2022.
\newblock Challenging big-bench tasks and whether chain-of-thought can solve
  them.
\newblock \emph{arXiv preprint arXiv:2210.09261}.

\bibitem[{Vendrow et~al.(2025)Vendrow, Vendrow, Beery, and
  Madry}]{vendrow2025platinum}
Joshua Vendrow, Edward Vendrow, Sara Beery, and Aleksander Madry. 2025.
\newblock \href {https://arxiv.org/abs/2502.03461} {Do large language model
  benchmarks test reliability?}
\newblock \emph{Preprint}, arXiv:2502.03461.

\bibitem[{Wang et~al.(2024)Wang, Ma, Feng, Zhang, Yang, Zhang, Chen, Tang,
  Chen, Lin et~al.}]{wang2024survey}
Lei Wang, Chen Ma, Xueyang Feng, Zeyu Zhang, Hao Yang, Jingsen Zhang, Zhiyuan
  Chen, Jiakai Tang, Xu~Chen, Yankai Lin, et~al. 2024.
\newblock A survey on large language model based autonomous agents.
\newblock \emph{Frontiers of Computer Science}, 18(6):186345.

\bibitem[{Wang et~al.(2025)Wang, Liu, Qian, Shen, Pan, Xu, Abbasi, Ji, and
  Zhang}]{wang2025context}
Rushi Wang, Jiateng Liu, Cheng Qian, Yifan Shen, Yanzhou Pan, Zhaozhuo Xu,
  Ahmed Abbasi, Heng Ji, and Denghui Zhang. 2025.
\newblock \href {https://doi.org/10.18653/v1/2025.emnlp-main.1003}
  {Rescorla-wagner steering of {LLM}s for undesired behaviors over
  disproportionate inappropriate context}.
\newblock In \emph{Proceedings of the 2025 Conference on Empirical Methods in
  Natural Language Processing}, pages 19810--19845, Suzhou, China. Association
  for Computational Linguistics.

\bibitem[{Wang et~al.(2023)Wang, Li, Wang, Bai, Luo, Zhang, Jojic, Xing, and
  Hu}]{wang2023promptagent}
Xinyuan Wang, Chenxi Li, Zhen Wang, Fan Bai, Haotian Luo, Jiayou Zhang, Nebojsa
  Jojic, Eric~P Xing, and Zhiting Hu. 2023.
\newblock Promptagent: Strategic planning with language models enables
  expert-level prompt optimization.
\newblock \emph{arXiv preprint arXiv:2310.16427}.

\bibitem[{Wei et~al.(2022)Wei, Wang, Schuurmans, Bosma, Ichter, Xia, Chi, Le,
  and Zhou}]{wei2022chain}
Jason Wei, Xuezhi Wang, Dale Schuurmans, Maarten Bosma, Brian Ichter, Fei Xia,
  Ed~H. Chi, Quoc~V. Le, and Denny Zhou. 2022.
\newblock \href {https://doi.org/10.52202/068431-1800} {Chain-of-thought
  prompting elicits reasoning in large language models}.
\newblock In \emph{Advances in Neural Information Processing Systems},
  volume~35, pages 24824--24837.

\bibitem[{Whitehead and Fit-Florea(2017)}]{whitehead2017floating}
Nathan Whitehead and Alex Fit-Florea. 2017.
\newblock Floating point and ieee-754 compliance for nvidia gpus.
\newblock \emph{Nvidia Whitepaper}.

\bibitem[{Xu et~al.(2022)Xu, Chen, Du, Shao, Wang, Li, and Yang}]{xu2022gps}
Hanwei Xu, Yujun Chen, Yulun Du, Nan Shao, Yanggang Wang, Haiyu Li, and Zhilin
  Yang. 2022.
\newblock Gps: Genetic prompt search for efficient few-shot learning.
\newblock \emph{arXiv preprint arXiv:2210.17041}.

\bibitem[{Xue et~al.(2025)Xue, Li, Liu, Wang, Shen, and Wang}]{xue2025towards}
Eric Xue, Yijiang Li, Haoyang Liu, Peiran Wang, Yifan Shen, and Haohan Wang.
  2025.
\newblock \href {https://doi.org/10.1609/aaai.v39i9.32978} {Towards
  adversarially robust dataset distillation by curvature regularization}.
\newblock In \emph{Proceedings of the AAAI Conference on Artificial
  Intelligence}, volume~39, pages 9041--9049.

\bibitem[{Yang et~al.(2023)Yang, Wang, Lu, Liu, Le, Zhou, and
  Chen}]{yang2023large}
Chengrun Yang, Xuezhi Wang, Yifeng Lu, Hanxiao Liu, Quoc~V Le, Denny Zhou, and
  Xinyun Chen. 2023.
\newblock Large language models as optimizers.
\newblock In \emph{The Twelfth International Conference on Learning
  Representations}.

\bibitem[{Yao et~al.(2023)Yao, Zhao, Yu, Du, Shafran, Narasimhan, and
  Cao}]{yao2023react}
Shunyu Yao, Jeffrey Zhao, Dian Yu, Nan Du, Izhak Shafran, Karthik Narasimhan,
  and Yuan Cao. 2023.
\newblock React: Synergizing reasoning and acting in language models.
\newblock In \emph{International Conference on Learning Representations
  (ICLR)}.

\bibitem[{Yu et~al.(2025)Yu, Yu, Wei, Luo, and Wang}]{yu2025sipdo}
Yaoning Yu, Ye~Yu, Kai Wei, Haojing Luo, and Haohan Wang. 2025.
\newblock Sipdo: Closed-loop prompt optimization via synthetic data feedback.
\newblock \emph{arXiv preprint arXiv:2505.19514}.

\bibitem[{Yuksekgonul et~al.(2025)Yuksekgonul, Bianchi, Boen, Liu, Lu, Huang,
  Guestrin, and Zou}]{yuksekgonul2024textgrad}
Mert Yuksekgonul, Federico Bianchi, Joseph Boen, Sheng Liu, Pan Lu, Zhi Huang,
  Carlos Guestrin, and James Zou. 2025.
\newblock \href {https://doi.org/10.1038/s41586-025-08661-4} {Optimizing
  generative {AI} by backpropagating language model feedback}.
\newblock \emph{Nature}, 639:609--616.

\bibitem[{Zhan et~al.(2024)Zhan, Xu, Tan, Song, and Xie}]{zhan2024cople}
Pengwei Zhan, Zhen Xu, Qian Tan, Jie Song, and Ru~Xie. 2024.
\newblock Unveiling the lexical sensitivity of llms: Combinatorial optimization
  for prompt enhancement.
\newblock \emph{arXiv preprint arXiv:2405.20701}.

\bibitem[{Zhang et~al.(2025{\natexlab{a}})Zhang, Wang, Ren, Jiang, Wang, and
  Liu}]{zhang2025ratt}
Jinghan Zhang, Xiting Wang, Weijieying Ren, Lu~Jiang, Dongjie Wang, and Kunpeng
  Liu. 2025{\natexlab{a}}.
\newblock Ratt: A thought structure for coherent and correct llm reasoning.
\newblock In \emph{Proceedings of the AAAI Conference on Artificial
  Intelligence}, volume~39, pages 26733--26741.

\bibitem[{Zhang et~al.(2024)Zhang, Ergen, Logeswaran, Lee, and
  Jurgens}]{zhang2024sprig}
Lechen Zhang, Tolga Ergen, Lajanugen Logeswaran, Moontae Lee, and David
  Jurgens. 2024.
\newblock Sprig: Improving large language model performance by system prompt
  optimization.
\newblock \emph{arXiv preprint arXiv:2410.14826}.

\bibitem[{Zhang et~al.(2025{\natexlab{b}})Zhang, Jin, Hu, Li, Kang, Luo, Song,
  and Wang}]{zhang2024revolve}
Peiyan Zhang, Haibo Jin, Leyang Hu, Xinnuo Li, Liying Kang, Man Luo, Yangqiu
  Song, and Haohan Wang. 2025{\natexlab{b}}.
\newblock \href {https://proceedings.mlr.press/v267/zhang25aj.html} {Revolve:
  Optimizing {AI} systems by tracking response evolution in textual
  optimization}.
\newblock In \emph{Proceedings of the 42nd International Conference on Machine
  Learning}, volume 267 of \emph{Proceedings of Machine Learning Research},
  pages 75216--75233. PMLR.

\bibitem[{Zhou et~al.(2023{\natexlab{a}})Zhou, Liu, Xu, Iyer, Sun, Mao, Ma,
  Efrat, Yu, Yu et~al.}]{zhou2023lima}
Chunting Zhou, Pengfei Liu, Puxin Xu, Srinivasan Iyer, Jiao Sun, Yuning Mao,
  Xuezhe Ma, Avia Efrat, Ping Yu, Lili Yu, et~al. 2023{\natexlab{a}}.
\newblock Lima: Less is more for alignment.
\newblock \emph{Advances in Neural Information Processing Systems},
  36:55006--55021.

\bibitem[{Zhou et~al.(2023{\natexlab{b}})Zhou, Wan, Vuli{\'c}, and
  Korhonen}]{CLAPS}
Han Zhou, Xingchen Wan, Ivan Vuli{\'c}, and Anna Korhonen. 2023{\natexlab{b}}.
\newblock Survival of the most influential prompts: Efficient black-box prompt
  search via clustering and pruning.
\newblock \emph{arXiv preprint arXiv:2310.12774}.

\bibitem[{Zhou et~al.(2022)Zhou, Muresanu, Han, Paster, Pitis, Chan, and
  Ba}]{APE}
Yongchao Zhou, Andrei~Ioan Muresanu, Ziwen Han, Keiran Paster, Silviu Pitis,
  Harris Chan, and Jimmy Ba. 2022.
\newblock Large language models are human-level prompt engineers.
\newblock In \emph{The eleventh international conference on learning
  representations}.

\end{thebibliography}

\appendix

\clearpage
\section{Related Work}
\label{related_work}

\paragraph{Automated Prompt Optimization.}
Manual prompt design, including chain-of-thought prompting \citep{wei2022chain}, can improve LLM performance but is difficult to scale across tasks. Automated approaches broadly include soft prompt tuning \citep{lestersoftprompt, hu2022lora,liu2024gpt}, which requires access to model parameters, and discrete prompt search, which can optimize closed-source models using textual \citep{wang2023promptagent, adv-ICL, sinha2024survival} or numerical \citep{APE, CLAPS, deng2022rlprompt, zhang2024sprig} feedback.

\paragraph{Textual Gradient-Based Learning.}
ProTeGi \citep{pryzant2023protegi} and TextGrad \citep{yuksekgonul2024textgrad} use natural-language critiques as gradients in the discrete space of text. An evaluator identifies flaws in a target model's output and proposes a semantic direction for editing the prompt. ProTeGi combines this signal with beam search and bandit selection, while REVOLVE \citep{zhang2024revolve} incorporates the historical evolution of responses. STEVE is complementary: it filters the examples from which gradients are generated and gates candidate updates against a preservation sample.

\paragraph{Quality of Learning Signals.}
A core challenge in iterative optimization is the quality of the guiding signal. Our Error-Driven Refinement mechanism is motivated by established principles in active learning and information theory \citep{nguyen2021information, li2024survey}. The central tenet of active learning \citep{settles2009active} is that a model learns most efficiently from examples it finds difficult or uncertain about. Correctly handled examples provide a low-information signal. Forcing an LLM to generate feedback on a correct output can lead to random or stylistic critiques that act as noise, degrading the prompt rather than improving it. This aligns with findings from instruction tuning: LIMA \citep{zhou2023lima} demonstrates that a small set of high-quality, diverse data is far more effective than a large volume of noisy or low-quality data. Related evidence from multimodal preference optimization shows that fine-grained, segment-level preference signals can outperform a standard, coarser objective on spatial reasoning tasks \citep{shen2025fine}.

\section{Experimental Setup Details}
\label{sec:appendix_exp_setup}

We will release the source code and evaluation data with the paper. This section provides additional experimental details.

\begin{table*}[h]
\caption{Summary of dataset statistics used in our experiments.}
\label{tab:dataset_details}
\centering
\small
\setlength{\tabcolsep}{4pt}
\begin{tabular}{lcccc}
\toprule
\textbf{Dataset} & \textbf{Train Set Size} & \textbf{Test Set Size} & \textbf{Preservation Sample Set Size} & \textbf{Candidate Size} \\
\midrule
\multicolumn{5}{l}{\textit{Mathematical Reasoning}} \\
GSM8k & 50 & 100 & 20 & 3 \\
MultiArith & 50 & 50 & 20 & 3 \\
\midrule
\multicolumn{5}{l}{\textit{Complex Commonsense Reasoning}} \\
StrategyQA & 50 & 100 & 20 & 3 \\
Navigate & 50 & 100 & 20 & 3 \\
\midrule
\multicolumn{5}{l}{\textit{Symbolic \& Procedural Reasoning}} \\
Object Counting & 50 & 100 & 20 & 3 \\
Penguins in a Table & 87 & 30 & 20 & 3 \\
Geometric Shapes & 50 & 100 & 20 & 3 \\
Date Understanding & 50 & 100 & 20 & 3 \\
\midrule
\multicolumn{5}{l}{\textit{Expert-Level Knowledge Reasoning}} \\
College Physics & 50 & 100 & 20 & 3 \\
Machine Learning & 67 & 23 & 20 & 3 \\
\bottomrule
\end{tabular}
\end{table*}

\subsection{Dataset Details}
\label{sec:appendix_datasets}

Table~\ref{tab:dataset_details} summarizes the original evaluation. ``Train Set" is the pool available to iterative optimization, from which $\mathcal{D}_{\text{hard}}$ and the eligible preservation pool are formed. At every iteration, $k=20$ preservation examples are freshly sampled from that fixed pool and shared across all candidate evaluations. A disjoint held-out validation set selects $p^*$; the test set is used only for final reporting. The modernized evaluation uses the complete 1,009-example GSM8K-Platinum set and 800 DS-1000 problems, alongside the BBEH Object Counting and Geometric Shapes tasks.

\subsection{Model and API Details}
\label{sec:appendix_model_details}

\begin{table*}[h]
\centering
\caption{Estimated token consumption and cost for a single STEVE optimization run on BBH Object Counting. Assumes $n=3$ candidates and a preservation set sample size of $k=20$. Costs are based on September 2025 pricing for \texttt{gpt-4o} (evaluator/optimizer) and \texttt{gpt-3.5-turbo-0125} (executor).}
\label{tab:cost_analysis}
\resizebox{\textwidth}{!}{
\begin{tabular}{lccc}
\toprule
\textbf{Task} & \textbf{Total API Calls} & \textbf{Est. Tokens} & \textbf{Est. Total Cost (USD)} \\
& \small{(Executor/Evaluator/Optimizer)} & \small{(Executor/Evaluator/Optimizer)}& \\
\midrule
BBH Object Counting (1 Run) & 2298 / 60 / 20 & $\sim$1,730,156 / $\sim$36,260 / $\sim$108,780 & $\sim$\$4.05 \\
\bottomrule
\end{tabular}
}
\end{table*}

All experiments were conducted using API access to the respective language models. The specific model versions used are as follows:
\begin{itemize}
    \item \textbf{Executor Model:} \texttt{gpt-3.5-turbo-0125}
    \item \textbf{Evaluator/Optimizer Models:} \texttt{gpt-4o} (version \texttt{gpt-4o-2024-05-13}), \texttt{gpt-5} (version \texttt{gpt-5-2025-08-07}), and \texttt{gemini-2.5-flash} (released June 17, 2025; exact snapshot unavailable).
    \item \textbf{Decoding Temperature:} 0.0
    \item \textbf{Top-p:} 0
    \item \textbf{Seed:} 42
    \item \textbf{Regularization Schedule:} $\lambda_t=1.5+0.1t$.

\end{itemize}
\subsection{Computational Cost Analysis}
\label{sec:cost_analysis}

The primary computational cost of iterative prompt optimization methods such as TextGrad and STEVE is the number of LLM API calls. Table~\ref{tab:cost_analysis} reports both calls and token usage.

Our method, STEVE, intentionally incurs a higher computational cost per iteration to ensure optimization stability. The additional cost arises from two main sources within our framework. First, instead of generating a single candidate, we generate $n$ diverse candidates to better explore the solution space. Second, and more significantly, our regularized verification step requires evaluating each of the $n$ candidates on a preservation set of size $k$. The dominant overhead of our method is therefore approximately $n \times (b+k)$ additional executor LLM calls per iteration compared to a non-verifying, single-candidate approach.

\subsection{Sensitivity to Holdout Size ($k$) and Cost-Performance Trade-off}
\label{sec:appendix_sensitivity_holdout}

In our implementation, we set the preservation sample size to $k=20$, an empirical choice that balances optimization stability against computational cost. Larger $k$ provides a stricter regression check but increases verification overhead. To quantify this trade-off, we ran an ablation on BBH Object Counting with $k\in\{5,20,40\}$.

\begin{table}[h]
\centering
\resizebox{\columnwidth}{!}{%
\begin{tabular}{llccc}
\toprule
\textbf{Dataset} & \textbf{Eval engine} & \textbf{Holdout size ($k$)} & \textbf{Best Acc (\%)} & \textbf{Avg cost / run (\$)} \\
\midrule
BBH Object Counting & GPT-4o & 5  & 92.0 & $\sim$2.0 \\
BBH Object Counting & GPT-4o & 20 & 95.7 & $\sim$2.2 \\
BBH Object Counting & GPT-4o & 40 & 91.0 & $\sim$2.7 \\
\bottomrule
\end{tabular}%
}
\caption{Ablation study on the holdout size ($k$) and its impact on performance and computational cost for the BBH Object Counting task.}
\label{tab:holdout_sensitivity}
\end{table}

\subsection{Sensitivity to the Regularization Schedule}
\label{sec:appendix_sensitivity_lambda}

We separately vary the initial value $\lambda_0$ and per-step increment $\Delta\lambda$ on BBH Object Counting, using \texttt{gpt-3.5-turbo-0125} as executor and \texttt{gpt-4o} as evaluator/optimizer. The default linear schedule, $\lambda_t=1.5+0.1t$, performs best among the tested settings. A very small increment provides weaker protection against late-stage over-specialization, whereas an increment of 1.0 becomes too conservative.

\begin{table}[h]
\centering
\small
\begin{tabular}{llc}
\toprule
\textbf{Parameter varied} & \textbf{Value} & \textbf{Accuracy (\%)} \\
\midrule
$\Delta\lambda$ ($\lambda_0=1.5$) & 0.01 & 94.3 \\
 & \textbf{0.1} & \textbf{95.7} \\
 & 1.0 & 88.3 \\
\midrule
$\lambda_0$ ($\Delta\lambda=0.1$) & 1.0 & 89.3 \\
 & \textbf{1.5} & \textbf{95.7} \\
 & 2.0 & 93.6 \\
\bottomrule
\end{tabular}
\caption{Sensitivity of STEVE to the linear regularization schedule on BBH Object Counting.}
\label{tab:lambda_sensitivity}
\end{table}
\section{Additional Comparisons with Non-Gradient Prompt Optimization Method}
\label{sec:appendix_evoprompt_comparison}

To provide broader comparisons beyond textual gradient-based methods, we additionally compare \textsc{STEVE} against EvoPrompt \citep{guo2023connecting}, a representative non-gradient evolutionary prompt optimizer. We evaluate both methods on several BigBench Hard (BBH) tasks under the same backbone model (GPT-4o) to ensure a fair, apples-to-apples comparison.

As shown in Table~\ref{tab:evoprompt_comparison}, \textsc{STEVE} is highly competitive with EvoPrompt and demonstrates consistent improvements over it across multiple tasks, with particularly strong gains in reasoning-heavy tasks such as ``Penguins in a Table'' and ``Object Counting''.

\begin{table}[h]
\centering
\resizebox{\columnwidth}{!}{%
\begin{tabular}{lcccc}
\toprule
\textbf{BBH Task} & \textbf{Base Prompt} & \textbf{EvoPrompt} & \textbf{Ours (STEVE)} & $\bm{\Delta}$ \\
\midrule
Navigate            & 60.1 & 94.2 & 95.6 & +1.4 \\
Object\_counting    & 77.9 & 87.6 & 95.7 & +8.1 \\
Penguins in a Table & 80.8 & 84.3 & 96.5 & +12.2 \\
Geometric Shapes    & 39.4 & 60.2 & 62.9 & +2.7 \\
\bottomrule
\end{tabular}%
}
\caption{Performance comparison between the Base Prompt, EvoPrompt, and \textsc{STEVE} on select BBH tasks using GPT-4o. The $\Delta$ column indicates the absolute performance improvement of \textsc{STEVE} over EvoPrompt.}
\label{tab:evoprompt_comparison}
\end{table}

\section{Open-Source Optimizer LLM}
\label{sec:appendix_opensource_llm}

We additionally tested smaller open-source models (Qwen3-VL-8B, DeepSeek-V3.2) as the optimizer/evaluator. Results below report best accuracy on each benchmark.

\begin{table}[h]
\centering
\resizebox{\columnwidth}{!}{%
\begin{tabular}{llc}
\toprule
\textbf{Dataset} & \textbf{Eval engine} & \textbf{Ours (STEVE)} \\
\midrule
StrategyQA          & Qwen3-VL-8B   & 91.0 \\
StrategyQA          & DeepSeek-V3.2 & 94.0 \\
BBH Object Counting & Qwen3-VL-8B   & 88.0 \\
BBH Object Counting & DeepSeek-V3.2 & 87.0 \\
GSM8K               & Qwen3-VL-8B   & 81.0 \\
GSM8K               & DeepSeek-V3.2 & 82.0 \\
\bottomrule
\end{tabular}%
}
\caption{Performance of \textsc{STEVE} using smaller open-source models as the optimizer and evaluator.}
\label{tab:opensource_optimizer}
\end{table}
\section{Implementation Details of STEVE}
\label{sec:appendix_implementation}

\subsection{Guidance for Selecting Evaluator/Optimizer Models}
We recommend a practical selection/validation protocol for new tasks:
\begin{itemize}
    \item \textbf{Evaluator faithfulness check:} on a small batch, verify that critiques reliably identify true failure causes rather than generic advice.
    \item \textbf{One-step sanity test:} run a single optimization iteration and measure whether candidate prompts improve the hard set without causing disproportionate regression on the verification set.
    \item \textbf{Stability check across seeds:} estimate variance over 2--3 short runs to detect unstable evaluator/optimizer combinations early.
\end{itemize}

STEVE is compatible with different model families, but users should validate the evaluator/optimizer using these lightweight diagnostics before committing to full runs.

\subsection{Prompts for Evaluator and Optimizer Models}
\label{sec:appendix_prompts}

\begin{tcolorbox}[colback=gray!5!white,colframe=gray!50!black,
  colbacktitle=gray!75!black,title=Evaluator Prompt, breakable]
``\textless OBJECTIVE\_FUNCTION\textgreater Your goal is to give feedback and criticism to the variable given the above evaluation output." \\
``Our only goal is to improve the above metric, and nothing else. \textless/OBJECTIVE\_FUNCTION\textgreater" \\
``This conversation is part of a larger system. The \textless OUTPUT\_OF\_FUNCTION\textgreater was later used as \{response\_desc\}." \\
``\textless OBJECTIVE\_FUNCTION\textgreater Your goal is to give feedback to the variable to address the following feedback on the OUTPUT\_OF\_FUNCTION: \{response\_gradient\} \textless/OBJECTIVE\_FUNCTION\textgreater" \\
``We are interested in giving feedback to the \{variable\_desc\} " \\
``for this conversation. Specifically, give feedback to the following span " \\
``of text: \textless VARIABLE\textgreater " \\
``\{variable\_short\} \textless/VARIABLE\textgreater" \\
``Given the above history, describe how the \{variable\_desc\} " \\
``could be improved to improve the \textless OBJECTIVE\_FUNCTION\textgreater. Be very creative, critical, and intelligent."
\end{tcolorbox}

The reproducibility of our method relies on the meta-prompts used to guide the evaluator and optimizer models. Above and below are the prompts used in our experiments.

\begin{tcolorbox}[colback=gray!5!white,colframe=gray!50!black,
  colbacktitle=gray!75!black,title=Optimizer Prompt]
\small
``Here is the role of the variable you will improve: \textless ROLE\textgreater\{variable\_desc\}\textless/ROLE\textgreater.''

``The variable is the text within the following span: \textless VARIABLE\textgreater \{variable\_short\} \textless/VARIABLE\textgreater''

``Here is the context and feedback we got for the variable:''

\textless CONTEXT\textgreater\{variable\_grad\}\textless/CONTEXT\textgreater

``Improve the variable (\{variable\_desc\}) using the feedback provided in \textless FEEDBACK\textgreater tags.''

``Send the improved variable '' \\
``in the following format:''

\{new\_variable\_start\_tag\}\{\{the improved variable\}\}\{new\_variable\_end\_tag\}

``Send ONLY the improved variable between the \textless IMPROVABLE\textgreater tags, and nothing else.'' 
\end{tcolorbox}

\subsection{Initial Prompts ($P_0$)}
\label{sec:appendix_initial_prompts}

All iterative optimization methods in our experiments began from a general Zero-shot Chain-of-Thought (CoT) prompt, $P_0$. This ensures that performance gains are a direct result of the optimization process. To accommodate specific output formats required by certain benchmarks, minor instructional text was added to a base prompt. Table~\ref{tab:initial_prompts_detailed} details the exact initial prompt used for each of the 10 datasets. No other task-specific modifications or in-context examples were included.

\section{A Qualitative Example of Over-specialization on Hard Cases}
Below is the final prompt optimized by hard cases without Regularized Verification on dataset StrategyQA. The prompt includes details about processes to solve specific hard examples.
\label{sec:qualilative_example}
\begin{tcolorbox}[colback=gray!5!white,colframe=gray!50!black,
  colbacktitle=gray!75!black,title= Prompt on Hard Cases after 12 Steps Training, breakable]
"Answer the following yes/no question. Begin with a clear 'Answer: True' or 'Answer: False' statement. First, identify the core question and distinguish between primary and secondary information to understand the specific context and intent. Ensure the initial answer is logically consistent with the provided data by performing a preliminary check. Pay special attention to key phrases or terms that might indicate specific contexts or conditions, such as dates and ages, to enhance contextual understanding. Define what constitutes a \"project\" for each entity, specifying categories such as TV shows, movies, specials, and spinoffs, and apply these definitions consistently. Evaluate the temporal context by assessing the timeline of events and their relevance to the current year or the year in question. Cross-reference multiple reliable sources for fact verification, listing potential sources and checking for consistency before finalizing the answer. Use a secondary model or external knowledge base for confirmation when necessary. Ensure all information directly contributes to the conclusion, explicitly stating how each piece supports the boolean answer. Structure the explanation in a step-by-step manner, ensuring each point logically leads to the conclusion. Recognize and address any hypothetical scenarios or specific conditions by identifying keywords or phrases that indicate such situations. Break down the question into logical components and evaluate each against the given conditions to determine the boolean outcome. Consider potential edge cases and how they might affect the outcome. Use precise language and clarify any terms that could be interpreted in multiple ways. Incorporate strategies for handling ambiguous queries, such as identifying potential ambiguities, seeking clarification, or providing a probabilistic answer when certainty is not achievable. Implement a confidence scoring mechanism to express certainty in the answer, providing a probability score or a statement of uncertainty when not fully confident, prompting further verification or clarification. Be aware of common biases and heuristics, critically evaluating their applicability to the specific case and considering exceptions to general rules. Reflect on training data to recall previous similar questions and their resolutions, applying learned patterns to new queries. After formulating an initial response, verify the answer by cross-referencing with a reliable knowledge base to ensure accuracy. Perform a self-assessment by reflecting on potential errors or misinterpretations, and adjust the response accordingly. Ensure alignment with the ground truth by checking the conclusion against a known correct answer or reliable source. Provide a clear and robust justification for the answer, exploring both direct and indirect factors. Use definitive language to avoid ambiguity and ensure logical consistency throughout the explanation. Focus on the specific query, filtering out extraneous details and concentrating on elements crucial to answering the question accurately. Use the following refined example as a reference for structuring your response: Given the conditions that [condition], the answer is [True/False] because [reasoning]. Incorporate an iterative feedback and learning loop to refine understanding and improve accuracy over time, analyzing incorrect answers to identify errors and adjust strategies accordingly."
\end{tcolorbox}

\section{A Qualitative Analysis of Successful STEVE-Optimized Prompts
}
\label{sec:qualilative_successful_example}
Below is an example of a STEVE-optimized prompt for BBH Object Counting tasks. The prompt excerpt is truncated for brevity.
\begin{tcolorbox}[colback=gray!5!white,colframe=gray!50!black,
  colbacktitle=gray!75!black,title=Successful STEVE-optimized prompts]
You will answer a reasoning question. Follow these steps to ensure accuracy and clarity:\\[0.5em]
1. Explicit Task Understanding: Restate the task in your own words to confirm understanding before proceeding.\\
2. Explicit Task Clarification: Identify, list, and sum all relevant numerical values from the input. Define task-specific terms (e.g., what constitutes a "vegetable") to avoid misclassification.\\
3. Enhanced Contextual Awareness: Maintain a running list of all relevant items or steps, ensuring consistent reference to avoid omissions or miscounts.\\
4. Enhanced Number Extraction Instructions: Extract and list all numbers from the input before performing any operations.\\
5. Detailed Summation Process: Perform the summation step-by-step, showing intermediate calculations for transparency and accuracy.\\
6. Explicit Error Analysis and Correction: List assumptions and check them against rules/common knowledge; provide a rationale for the answer.\\
...
\end{tcolorbox}

\section{LLM Usage}
We utilized an LLM solely for the purpose of refining the prose and enhancing the clarity of this paper. The model was prompted to correct grammatical errors, improve sentence structure, and polish writing. All intellectual contributions, including the core ideas, experimental design, and analysis, are exclusively the work of the authors. The LLM's role was strictly limited to that of a writing aid and did not contribute to the scientific content of this research.

\begin{table*}[h!]
\centering
\caption{Initial prompts ($P_0$) used for each benchmark in our experiments.}
\label{tab:initial_prompts_detailed}
\begin{tabular}{ll}
\toprule
\textbf{Dataset} & \textbf{Initial Prompt ($P_0$) Text} \\
\midrule
\multicolumn{2}{l}{\textit{Mathematical Reasoning}} \\
GSM8k & \parbox[t]{0.7\linewidth}{\texttt{You will answer a reasoning question. Think step by step. \\ The last line of your response should be of the following format: \\ 'Answer: \$VALUE' where VALUE is a numerical value.}} \\
\addlinespace[1em]
MultiArith & \parbox[t]{0.7\linewidth}{\texttt{You will solve arithmetic word problems. Think step by step and output your final answer in the format 'Answer: \$NUMBER'.}} \\
\midrule
\multicolumn{2}{l}{\textit{Complex Commonsense Reasoning}} \\
StrategyQA & \parbox{0.74\textwidth}{\raggedright\texttt{Answer the following yes/no question. Think step by step and provide reasoning before answering. The last line of your response should be of the following format: 'Answer: True' or 'Answer: False'.}} \\
\addlinespace[1em]
Navigate & \parbox{0.74\textwidth}{\raggedright\texttt{You will answer a reasoning question. Think step by step. The last line of your response should be of the following format: 'Answer: \$VALUE' where VALUE is a numerical value.}} \\
\midrule
\multicolumn{2}{l}{\textit{Symbolic \& Procedural Reasoning}} \\
Object Counting & \parbox[t]{0.7\linewidth}{\texttt{You will answer a reasoning question. Think step by step. \\ The last line of your response should be of the following format: \\ 'Answer: \$VALUE' where VALUE is a numerical value.}} \\
\addlinespace[1em]
Penguins in a Table & \parbox[t]{0.7\linewidth}{\texttt{You will answer a reasoning question. Think step by step. \\ The last line of your response should be of the following format: \\ 'Answer: \$VALUE' where VALUE is a numerical value.}} \\
\addlinespace[1em]
Geometric Shapes & \parbox[t]{0.7\linewidth}{\texttt{You will answer a reasoning question. Think step by step. \\ The last line of your response should be of the following format: \\ 'Answer: \$VALUE' where VALUE is a numerical value.}} \\
\addlinespace[1em]
Date Understanding & \parbox[t]{0.7\linewidth}{\texttt{Answer the following multiple choice question. Think step by step. \\ The last line must be 'Answer: \$LETTER'. LETTER must be one of A, B, C, D, E, or F.}} \\
\midrule
\multicolumn{2}{l}{\textit{Expert-Level Knowledge Reasoning}} \\
College Physics & \parbox{0.74\textwidth}{\raggedright\texttt{You will answer multiple-choice questions. Think step by step. The goal is to select the correct final answer from the choices.}} \\
\addlinespace[1em]
Machine Learning & \parbox{0.74\textwidth}{\raggedright\texttt{You will answer multiple-choice questions. Think step by step. The goal is to select the correct final answer from the choices.}} \\
\bottomrule
\end{tabular}
\end{table*}

\end{document}